\documentclass{article}

\usepackage[preprint]{corl_2022} 

\usepackage{graphics}
\usepackage{epsfig} 
\usepackage{mathptmx}
\usepackage{times} 

\usepackage{mathtools}
\usepackage{amsmath}
\usepackage{amssymb}
\usepackage{bm}
\usepackage{eucal}

\newcommand{\bx}{\mathbf{x}}

\newcommand{\bu}{\mathbf{u}}
\newcommand{\bp}{\mathbf{p}}
\newcommand{\bs}{\mathbf{s}}
\newcommand{\bsdot}{\dot{\bs}}
\newcommand{\bsddot}{\ddot{\bs}}

\newcommand{\jointTau}{\boldsymbol{\tau}}
\newcommand{\jointPos}{\mathbf{q}}
\newcommand{\jointVel}{\dot{\jointPos}}
\newcommand{\jointAcc}{\ddot{\jointPos}}

\newcommand{\linVel}{\dot{\bp}}

\newcommand{\grf}{\mathbf{f}}
\newcommand{\GRF}{\mathbf{F}}

\newcommand{\R}{\mathbb{R}}

\newcommand{\targetTime}{\mathcal{T}}
\newcommand{\Forward}{\mathcal{F}}
\newcommand{\Dist}{\mathcal{D}}
\newcommand{\model}{\boldsymbol{\pi}}
\newcommand{\bZero}{\mathbf{0}}

\newcommand{\norm}[1]{ ||#1||} 
\newcommand{\normal}{\mathcal{N}}

\title{Learning Agile Paths from Optimal Control}

\author{
  Alex Beaudin\\
  McGill University\\
  Department of Computer Science\\
  Montreal, Quebec, Canada\\
  \texttt{alex.beaudin@mail.mcgill.ca} \\
  \And 
  Hsiu-Chin Lin\\
  McGill University \\
  Department of Computer Science\\
  Montreal, Quebec, Canada\\
  \texttt{hsiu-chin.lin@cs.mcgill.ca} \\
}

\begin{document}
\maketitle


\begin{abstract}
    Efficient motion planning algorithms are of central importance for deploying robots in the real world.
    Unfortunately, these algorithms often drastically reduce the dimensionality of the problem for the sake of feasibility, thereby foregoing optimal solutions.
    This limitation is most readily observed in agile robots, where the solution space can have multiple additional dimensions.
    Optimal control approaches partially solve this problem by finding optimal solutions without sacrificing the complexity of the environment, but do not meet the efficiency demands of real-world applications.
    This work proposes an approach to resolve these issues simultaneously by training a machine learning model on the outputs of an optimal control approach. 
\end{abstract}



\section{Introduction}
	
Autonomous robotic systems are of particular interest for many fields, especially those that can be dangerous for human intervention like search and rescue, and maintenance on rigs.
However, motion planning in unstructured environment is still a hard problem for legged robots and their success depends largely on their ability to plan their paths robustly.
Moreover, the method in which a controller deals with obstacles has great consequences on the planned trajectory, and these optimizations are quintessential in generating agile motions for real-world robots.

Trajectory optimization is a common practice for generating motion for legged systems~\cite{kalakrishnan2011stomp,posa2016optimization,bjelonic2020rolling}, since it can produce optimal trajectories which satisfy the physical and environmental constraints of the robot. 
However, the solution from trajectory optimization is only valid for a particular pair of initial and target positions, and one needs to re-plan if the pair changes. 
Due to high-dimensionality and complexity, solving such an optimization problem for legged robots is infeasible in real-time.

Previous work simplified the problem by using a reduced-order model~\cite{apgar2018fast} and refining the trajectory using model predictive control~\cite{mit-cheetah}.
However, the issue is exacerbated in the presence of obstacles, since collision avoidance constraints are non-linear algebraic constraints and so harder to solve.

In recent years, imitation learning~\cite{schaal.1996,duan2017one} and reinforcement learning~\cite{yang2022fast,MIT2019Atkeson} have become the dominant focus in the research community.
The data-driven approach offers a global solution and removes the hurdle of re-planning.
On the other hand, collecting data for imitation learning is labour intensive work, which can be done by using motion capture~\cite{lin2014novel} or using animal data~\cite{RoboImitationPeng20}, which is extremely difficult on legged robots. 
Reinforcement learning does not require any data, but it is extremely time-consuming to learn a policy.


For planning with obstacles, most work focuses on modelling the environment as a 2-dimensional grid that represents the height of the obstacles~\cite{lee2020learning}.
The collision avoidance method finds the traversable paths in the plane~\cite{gasparino2022wayfast}.
However, the paths may be sub-optimal, since completely circumventing an obstacle is time consuming at best, and completely impossible at worst.
  
To mitigate the limitations of optimal control and imitation learning, we propose a self-supervised learning approach for efficient 3D collision avoidance in real-time.
Specifically, we generate a set of motion data from optimal control with a reduced model to create a rough plan and learn a policy that reproduces the motion data.
The learned policy is refined through whole-body model predictive control which satisfies the physical constraints of the robot.

\section{Background}


Let $\bx_k,\bu_k$ represent the states and actions of the robot at time-step $k$, the goal of optimal control is to find a trajectory, a set of $\bx,\bu$, such that a given cost function is minimized. 
Assuming that $\bx^i$, $\bx^t$ are the initial and target state of the robot specified by the user, a typical problem can be formulated as the following
\begin{equation}
\begin{aligned}
    \min_{\bx_0,\dots\bx_N, \bu_0, \bu_N}  \quad   & \sum \mathcal{L}(\bx_k,\bu_k )    & \text{ cost function} \\
    \text{subject to} \quad         & \bx_0 = \bx^i                     & \text{ initial condition} \\
                                    & \bx_N = \bx^t                     & \text{ terminal condition} \\
                                    & \bx_{k+1} = \Forward(\bx_k, \bu_k)    & \text{ forward dynamics } \\
                                    & \vdots            &\text{other constraints}
\end{aligned}    
\end{equation}


For legged robots, motion planning is normally done through optimization. 
It is well-known that the states of legged robots drift, and predicting a long trajectory is not ideal.
In addition, trajectory optimization, especially for long horizon, is not feasible in  real-time. This is particularly an issue in the presence of obstacles, since collision constraints are generally non-linear and thus require non-linear solvers. 
Most people combine trajectory optimization for long horizon planning with model predictive control for short horizon planning real time planning and control.

In the proposed work, we will use trajectory optimization to plan a rough path for the robot torso while avoiding collisions with the environment. The outcome of trajectory optimization is generated using an approximated model, which may not be realistic for the robot. Therefore, we use model-predictive control to refine the path from the reduced model.

\section{Methods}
Let $\jointPos,\jointVel,\jointAcc \in\R^{12}$ denote the joint positions, velocities, and accelerations of a 12 degree-of-freedom quadruped robot.
We assume that the target position of the robot torso is given, and the environment constraints are fully provided.
Our goal is find the control torques $\jointTau\in\R^{12}$ that can reach the given target while avoiding the obstacles.
The objective of the proposed framework is to enable robots to learn a novel skill from self-labelled data. 

\subsection{Autonomous Data Generation}
\label{sec:optimization}
    Our approach is to formulate the problem as an optimal control problem.
    We simplify the problem by focusing on the trajectory of the robot torso. 
    \begin{equation}
        \bs = [x,y,z,\theta_z]^T \in \R^4
    \end{equation}
    where $x,y,z$ denote the translation of the torso, $\theta$ denotes the rotation about its local z-axis, and roll and pitch are fixed during the optimization.
    Given the initial position of the robot state $\bs^i$, 
    the task is to find the a sequence of states $\{\bs_k\}_{k=1}^N$ that guides the robot from its initial pose $\bs^i\in \R^4$ to its target pose $\bs^t \in \R^4$ while minimizing the time $\targetTime$ and avoiding the obstacles at position $\bs^o$.
    
    We formulate this as a trajectory optimization problem, where the state is the positions and velocities $\bx=\left[\bs,\bsdot\right]^T \in\R^8$, and the command is the acceleration $\bu=\bsddot \in \R^4$.
    The decision variables are the sequences of N states and commands, as described in Equation~\ref{equ:optimization}.
    
    
    
    \vspace{-2mm}
    \begin{equation}
        \begin{aligned}
            \min_{
                \bx_0, \cdots, \bx_N,
                \bu_0, \cdots, \bu_N,
            } & \qquad \targetTime & \text{ minimum time} \\
            \text{subject to } \qquad   & \bx_{k+1} = \Forward(\bx_k, \bu_k), \quad \forall k=0,\dots, N-1 & \text{ forward dynamics} \\
                                        &  \bx_0 = [\bs^i, \bZero]^T, \bx_1 = \bZero & \text{ initial condition}\\
                                        &  \bx_N = [\bs^t, \bZero]^T, \bu_N = \bZero & \text{ terminal condition}\\
                                        & \bx_{min} \leq \bx_k \leq \bx_{max}, \quad \forall k=0,\dots, N & \text{ state boundary conditions} \\
                                        & \bu_{min} \leq \bu_k \leq \bu_{max}, \quad \forall k=0,\dots, N & \text{ action boundary conditions} \\
                                        & \Dist (\bp_k, \bp^o) \geq \epsilon , \quad \forall k=0,\dots, N & \text{ collision constraints} \\
        \end{aligned}
        \label{equ:optimization}
    \end{equation}

    where 
    $\Forward$ defines the dynamic equation of the system, 
    $\bx_{min}$, $\bx_{max}$, $\bu_{min}$,$\bu_{max}$ are the lower and upper bounds of states and actions, 
    and $\Dist(\bp_k, \bp^o)$ denotes the distance between the robot and the obstacles.
    This problem is transcribed into a direct collocation problem~\cite{direct-collocation} and solved using CasADi~\cite{casadi}.

\subsection{Learning a Predictive Model}
\label{sec:learning}

Assume that data are generated using the formal section as a set of positions $\bp$ and velocities $\linVel$, our goal is to learn a mapping $\model(.) \in \R^4 \rightarrow \R^4$ that predicts the most suitable velocity given the current state $\tilde{\linVel}_k = \model(\bp_k)$).

We use a neural network to encode this relationship. 
The architecture consists of six fully connected layers, each separated by a $\tanh$ activation function.
The network is trained using stochastic gradient descent to optimize the mean squared error between the generated $\linVel_k$ and the predicted velocity $\tilde{\linVel}_k$.

\subsection{Whole-Body Model Predictive Control}
Assuming that 
$\bx_k^{ref}$ is the reference state produced by the learnt model in Sec.~\ref{sec:learning},
the whole-body model predictive control~\cite{di2018dynamic} computes the torques $\jointTau \in\R^{12}$ that track the desired inputs $\bx_k^{ref}$.
This component minimizes the ground reaction force $\GRF=[\grf_1, \dots, \grf_K]^T$ for $K$ stance legs while satisfying the physical constraints of the robot and the friction cone constraints, which prevent slippage. 

    \begin{equation}
    \begin{aligned}
        \min_{\GRF_1, \GRF_2, \dots, \GRF_M}         
                                & \sum \norm{\bx_k - \bx_k^{ref}} + \norm{\GRF_k}         & \text{ loss function} \\
        \text{subject to}\quad  & \bx_{k+1} = \Forward(\bx_{k},\bu_k)                   & \text{ forward dynamics}\\
                                & \mu \lambda_z \geq \sqrt{ \lambda_x^2+\lambda_y^2 }  & \text{ friction cone constraints} \\
                                & \jointTau^{min} \leq \jointTau_k \leq \jointTau^{max}               & \text{ torque limit constraints} \\
                                & \jointPos^{min} \leq \jointPos_k \leq \jointPos^{max}               & \text{ joint limit constraints} \\
    \end{aligned}
    \end{equation}
Here, the horizon $M$ is a relatively small number. Once the $\GRF$ is found, the first solution $\GRF_0$ is taken and the rest are discarded. The ground reaction forces are converted into the equivalent torques.

The swing leg motion is independent from the model predictive control. Given the desired base velocity, we use a simple footstep planner which reads the desired base velocity and generate the next footstep position~\cite{raibert1986legged}. A simple interpolation is applied between the current footstep position and the next footstep position. Then, this is tracked by standard PD control.

\begin{figure}[h!]
    \centering
  \includegraphics[width=0.85\textwidth]{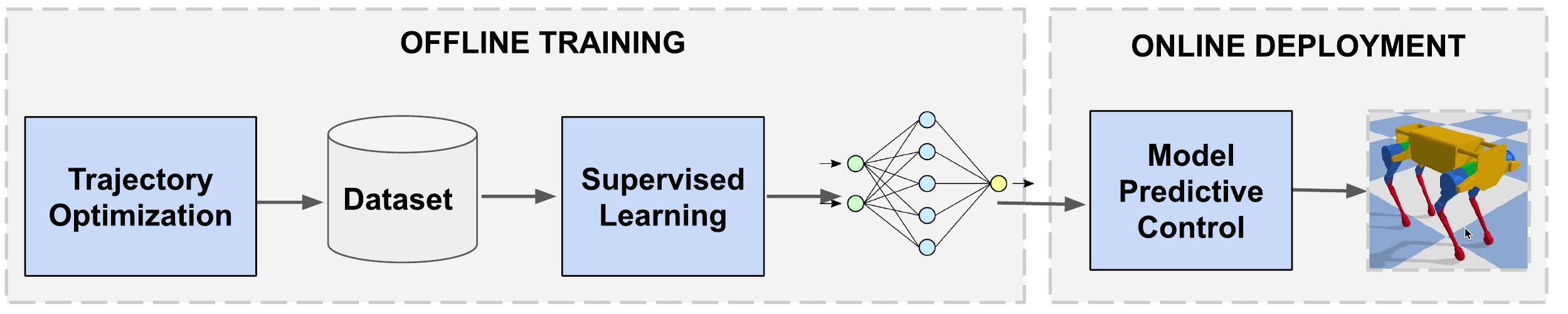}
  \caption{The pipeline of self-supervised collision avoidance planning}
  \label{fig:flowchart}
\end{figure}

Finally, a flowchart of the proposed framework is summarized in Fig.~\ref{fig:flowchart}.

\section{Experiments}
The experiments were carried out on a quadruped robot in PyBullet~\cite{PyBullet}. 
We created a simulated world where the robot needs to move from its initial position to its target position with an obstacle between them.
The nominal height of the robot torso is 28 cm, and the height of the table is 25 cm.
The robot can crawl under the obstacle only if it lowers its torso height.

\begin{figure}[ht!]
    \centering
    
    \includegraphics[height=0.2\textwidth]{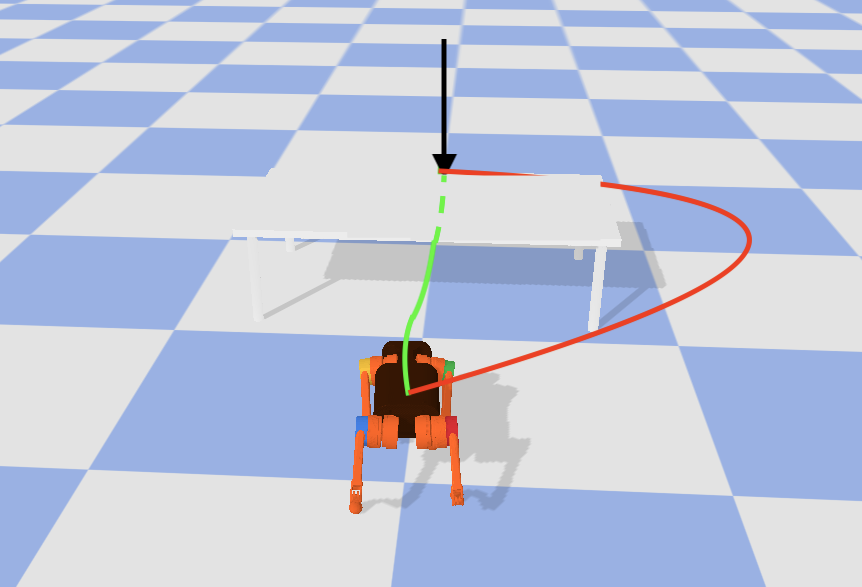}
    \includegraphics[height=0.2\textwidth]{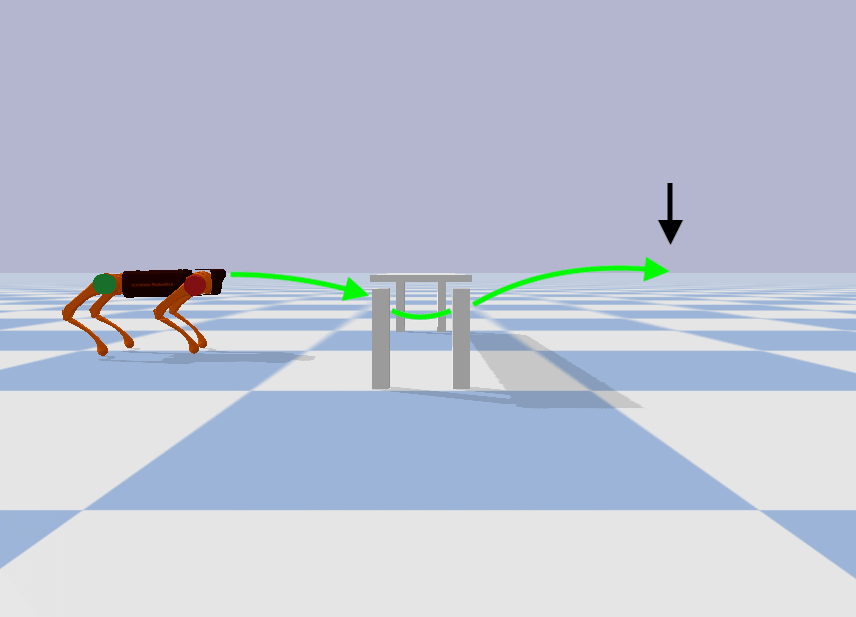}
  
    \caption{The experimental setup from the front and the side view. The robot starts at the origin and must move to the target (black arrow). The red curve shows the path generated by planning motion in 2D, and the green curve shows the path generated via 3D optimal control. }
  
  \label{fig:setup}
\end{figure}

Fig.~\ref{fig:setup} illustrates the simulated setup.
The robot starts at the origin and must move to the black arrow at $(3, 0)$. The red curve shows the path generated by planning the 2D motion, and the green curve shows the path generated by the 3D optimal control approach.

The target position is $(3,0)$, the table is placed at $(1.5,0)$, and the initial positions of the robot are randomly drawn from {$\bp^i \sim \normal([0.5, 0.066, 0.026], [0.5,0.66,0.02])$}.
We use the trajectory optimization method discussed in Sec.~\ref{sec:optimization} to generate a path for each initial position, which yields 10000 trajectories, each with $
\approx 100$ data points.



We use the methods from Sec.~\ref{sec:learning} for learning a predictive model.
The network architecture is [256,1024,1024,1024,1024,256] in the hidden layers, 
and it took 80 seconds to train a model. This was done with batch sizes of 1024 data points for 20 epochs with the stochastic gradient descent optimizer and an initial learning rate of 0.5. The train-validate-test size proportions were $80\%-10\%-10\%$.
The results of the model can achieve average mean squared error of $10^{-5}$.

\begin{figure}[h!]
    \includegraphics[width=0.245\textwidth]{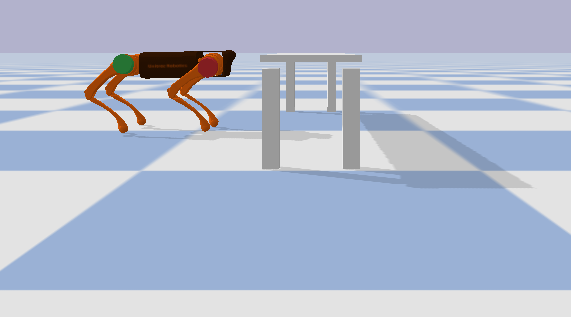}
    \includegraphics[width=0.245\textwidth]{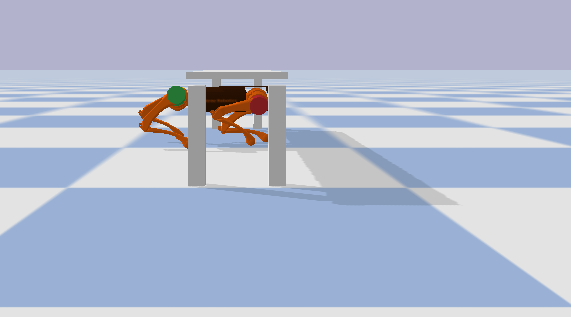}
    \includegraphics[width=0.245\textwidth]{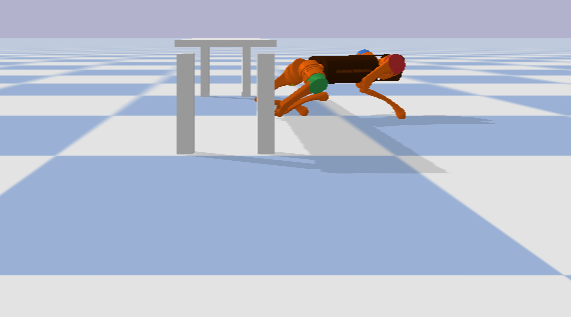}
    \includegraphics[width=0.245\textwidth]{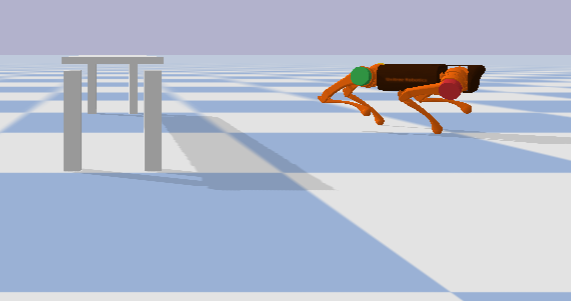}
    \caption{A snapshot of motion generated from learned model}
  \label{fig:results}
\end{figure}

    
Fig.~\ref{fig:results} shows the snapshot of an example trajectory generated using the proposed method. 
We can see that the robot lower is body in order to crawl under the table. 
    


\section{Conclusion}
\label{sec:conclusion}

This work proposes a self-supervised learning approach to learn a rough plan for a quadruped robot maneuver around obstacles.
We use optimal control to generate a rough plan and then use supervised learning to learn a predictive model.
The learned model provides the desired base motion and then it is refined using model predictive control for whole-body control.
Further improvements include relaxing more control variables to include the pitch and roll of the base and incorporating cameras and LiDARs for perceiving the environment.

    



\clearpage
\acknowledgments{We acknowledge the support of the Natural Sciences and Engineering Research Council of Canada (NSERC).\\
Nous remercions le Conseil de recherches en sciences naturelles et en génie du Canada (CRSNG) de son soutien.}


\bibliography{example}  

\begin{thebibliography}{18}
\providecommand{\natexlab}[1]{#1}
\providecommand{\url}[1]{\texttt{#1}}
\expandafter\ifx\csname urlstyle\endcsname\relax
  \providecommand{\doi}[1]{doi: #1}\else
  \providecommand{\doi}{doi: \begingroup \urlstyle{rm}\Url}\fi

\bibitem[Kalakrishnan et~al.(2011)Kalakrishnan, Chitta, Theodorou, Pastor, and
  Schaal]{kalakrishnan2011stomp}
M.~Kalakrishnan, S.~Chitta, E.~Theodorou, P.~Pastor, and S.~Schaal.
\newblock Stomp: Stochastic trajectory optimization for motion planning.
\newblock In \emph{IEEE international conference on robotics and automation},
  pages 4569--4574, 2011.

\bibitem[Posa et~al.(2016)Posa, Kuindersma, and Tedrake]{posa2016optimization}
M.~Posa, S.~Kuindersma, and R.~Tedrake.
\newblock Optimization and stabilization of trajectories for constrained
  dynamical systems.
\newblock In \emph{IEEE International Conference on Robotics and Automation
  (ICRA)}, pages 1366--1373, 2016.

\bibitem[Bjelonic et~al.(2020)Bjelonic, Sankar, Bellicoso, Vallery, and
  Hutter]{bjelonic2020rolling}
M.~Bjelonic, P.~K. Sankar, C.~D. Bellicoso, H.~Vallery, and M.~Hutter.
\newblock Rolling in the deep--hybrid locomotion for wheeled-legged robots
  using online trajectory optimization.
\newblock \emph{IEEE Robotics and Automation Letters}, 5\penalty0 (2):\penalty0
  3626--3633, 2020.

\bibitem[Apgar et~al.(2018)Apgar, Clary, Green, Fern, and Hurst]{apgar2018fast}
T.~Apgar, P.~Clary, K.~Green, A.~Fern, and J.~W. Hurst.
\newblock Fast online trajectory optimization for the bipedal robot cassie.
\newblock In \emph{Robotics: Science and Systems}, volume 101, page~14, 2018.

\bibitem[Di~Carlo et~al.(2018)Di~Carlo, Wensing, Katz, Bledt, and
  Kim]{mit-cheetah}
J.~Di~Carlo, P.~M. Wensing, B.~Katz, G.~Bledt, and S.~Kim.
\newblock Dynamic locomotion in the mit cheetah 3 through convex
  model-predictive control.
\newblock In \emph{IEEE/RSJ international conference on intelligent robots and
  systems (IROS)}, pages 1--9, 2018.

\bibitem[Schaal(1996)]{schaal.1996}
S.~Schaal.
\newblock Learning from demonstration.
\newblock \emph{Advances in neural information processing systems}, 9, 1996.

\bibitem[Duan et~al.()Duan, Andrychowicz, Stadie, Jonathan~Ho, Schneider,
  Sutskever, Abbeel, and Zaremba]{duan2017one}
Y.~Duan, M.~Andrychowicz, B.~Stadie, O.~Jonathan~Ho, J.~Schneider,
  I.~Sutskever, P.~Abbeel, and W.~Zaremba.
\newblock One-shot imitation learning.
\newblock \emph{Advances in neural information processing systems}, 30.

\bibitem[Yang et~al.(2022)Yang, Zhang, Coumans, Tan, and Boots]{yang2022fast}
Y.~Yang, T.~Zhang, E.~Coumans, J.~Tan, and B.~Boots.
\newblock Fast and efficient locomotion via learned gait transitions.
\newblock In \emph{Conference on Robot Learning}, pages 773--783, 2022.

\bibitem[Li et~al.(2019)Li, Geyer, Atkeson, and Rai]{MIT2019Atkeson}
T.~Li, H.~Geyer, C.~G. Atkeson, and A.~Rai.
\newblock Using deep reinforcement learning to learn high-level policies on the
  atrias biped.
\newblock In \emph{2019 International Conference on Robotics and Automation
  (ICRA)}, pages 263--269, 2019.
\newblock \doi{10.1109/ICRA.2019.8793864}.

\bibitem[Lin et~al.(2014)Lin, Howard, and Vijayakumar]{lin2014novel}
H.-C. Lin, M.~Howard, and S.~Vijayakumar.
\newblock A novel approach for representing and generalising periodic gaits.
\newblock \emph{Robotica}, 32\penalty0 (8):\penalty0 1225--1244, 2014.

\bibitem[Peng et~al.(2020)Peng, Coumans, Zhang, Lee, Tan, and
  Levine]{RoboImitationPeng20}
X.~B. Peng, E.~Coumans, T.~Zhang, T.-W.~E. Lee, J.~Tan, and S.~Levine.
\newblock Learning agile robotic locomotion skills by imitating animals.
\newblock In \emph{Robotics: Science and Systems}, 2020.

\bibitem[Lee et~al.(2020)Lee, Hwangbo, Wellhausen, Koltun, and
  Hutter]{lee2020learning}
J.~Lee, J.~Hwangbo, L.~Wellhausen, V.~Koltun, and M.~Hutter.
\newblock Learning quadrupedal locomotion over challenging terrain.
\newblock \emph{Science robotics}, 5\penalty0 (47):\penalty0 eabc5986, 2020.

\bibitem[Gasparino et~al.(2022)Gasparino, Sivakumar, Liu, Velasquez, Higuti,
  Rogers, Tran, and Chowdhary]{gasparino2022wayfast}
M.~V. Gasparino, A.~N. Sivakumar, Y.~Liu, A.~E. Velasquez, V.~A. Higuti,
  J.~Rogers, H.~Tran, and G.~Chowdhary.
\newblock Wayfast: Navigation with predictive traversability in the field.
\newblock \emph{IEEE Robotics and Automation Letters}, 7\penalty0 (4):\penalty0
  10651--10658, 2022.

\bibitem[Von~Stryk(1993)]{direct-collocation}
O.~Von~Stryk.
\newblock \emph{Numerical solution of optimal control problems by direct
  collocation}.
\newblock Springer, 1993.

\bibitem[Andersson et~al.(2019)Andersson, Gillis, Horn, Rawlings, and
  Diehl]{casadi}
J.~A. Andersson, J.~Gillis, G.~Horn, J.~B. Rawlings, and M.~Diehl.
\newblock Casadi: a software framework for nonlinear optimization and optimal
  control.
\newblock \emph{Mathematical Programming Computation}, 11\penalty0
  (1):\penalty0 1--36, 2019.

\bibitem[Di~Carlo et~al.(2018)Di~Carlo, Wensing, Katz, Bledt, and
  Kim]{di2018dynamic}
J.~Di~Carlo, P.~M. Wensing, B.~Katz, G.~Bledt, and S.~Kim.
\newblock Dynamic locomotion in the mit cheetah 3 through convex
  model-predictive control.
\newblock In \emph{IEEE/RSJ international conference on intelligent robots and
  systems (IROS)}, pages 1--9, 2018.

\bibitem[Raibert(1986)]{raibert1986legged}
M.~H. Raibert.
\newblock \emph{Legged robots that balance}.
\newblock MIT press, 1986.

\bibitem[Coumans and Bai(2016)]{PyBullet}
E.~Coumans and Y.~Bai.
\newblock Pybullet, a python module for physics simulation for games, robotics
  and machine learning, 2016.

\end{thebibliography}

\end{document}